\documentclass[lettersize,journal]{IEEEtran}
\usepackage{amsmath,amsfonts}
\usepackage{array}
\usepackage[caption=false,font=normalsize,labelfont=sf,textfont=sf]{subfig}
\usepackage{textcomp}
\usepackage{stfloats}
\usepackage{url}
\usepackage{verbatim}
\usepackage{graphicx}
\usepackage{cite}
\usepackage[ruled]{algorithm2e}
\usepackage{multirow}
\usepackage{pifont}
\usepackage{bm}
\usepackage{booktabs}
\usepackage{colortbl}
\definecolor{mypink}{rgb}{.99,.91,.95}
\definecolor{mygray}{gray}{.9}
\begin{document}

\title{Visible-Infrared Person Re-Identification via Spectral-Aware Softmax }

\author{Lei Tan, Pingyang Dai, Qixiang Ye, ~\IEEEmembership{Senior Member,~IEEE}, Mingliang Xu, ~\IEEEmembership{Member,~IEEE}, Yongjian Wu, Rongrong Ji, ~\IEEEmembership{Senior Member,~IEEE}
\thanks{Lei Tan, Pingyang Dai and Rongrong Ji are with the Media Analytics and Computing Laboratory, Department of Artificial Intelligence,
School of Informatics, and also with Institute of Artificial Intelligence, and Fujian Engineering Research Center of Trusted Artificial Intelligence Analysis and Application, Xiamen University, 361005, China. (e-mail:
tanlei@stu.xmu.edu.cn; pydai@xmu.edu.cn; rrji@xmu.edu.cn).}

\thanks{Qixiang Ye is with the Peng Cheng Laboratory, Shenzhen 518066, China,
and also with the School of Electronics, Electrical and Communication
Engineering, University of Chinese Academy of Sciences, Beijing 100049,
China (e-mail: qxye@ucas.ac.cn).}

\thanks{Mingliang Xu is with the School of Information Engineering, Zhengzhou
University, Zhengzhou 450000, China (e-mail: iexumingliang@zzu.edu.cn).}

\thanks{Yongjian Wu is with the Youtu Laboratory,
Tencent Technology (Shanghai) Co. Ltd, Shanghai 361005, China (e-mail:
littlekenwu@tencent.com;).}}
\markboth{Journal of \LaTeX\ Class Files,~Vol.~14, No.~8, August~2021}%
{Shell \MakeLowercase{\textit{et al.}}: A Sample Article Using IEEEtran.cls for IEEE Journals}


\maketitle

\begin{abstract}
Visible-infrared person re-identification (VI-ReID) aims to match specific pedestrian images from different modalities. 
Although suffering an extra modality discrepancy, existing methods still follow the softmax loss training paradigm, which is widely used in single-modality classification tasks. 
The softmax loss lacks an explicit penalty for the apparent modality gap, which adversely limits the performance upper bound of the VI-ReID task.
In this paper, we propose the spectral-aware softmax (SA-Softmax) loss, which can fully explore the embedding space with modality information and has clear interpretability. 
Specifically, SA-Softmax loss utilizes an asynchronous optimization strategy based on the modality prototype instead of the synchronous optimization based on the identity prototype in the original softmax loss.
To encourage high overlapping between two modalities, SA-Softmax optimizes each sample by the prototype from another spectrum.
Based on the observation and analysis of SA-Softmax, we modify the SA-Softmax by using the Feature Mask and Absolute-Similarity Term to alleviate the ambiguous optimization during model training.
Extensive experimental evaluations on RegDB and SYSU-MM01 demonstrate the superior performance of SA-Softmax in cross-modality conditions.

\end{abstract}
\section{Introduction}

Person re-identification (Re-ID) which aims at retrieving a specific person over a distributed set of non-overlapping cameras has made unprecedented progress~\cite{eom2019learning,Zhai2020ad,zheng2019pyramidal,zheng2019joint}.
Various efforts have been made to solve posture variations, resolution, occlusion, and blur under visible conditions. 
Nevertheless, most surveillance systems use near-infrared (NIR) images instead of visible (VIS) images under low-light conditions, as the near-infrared cameras provide an efficient way to obtain high-quality images under poor illumination. 
The problem of visible infrared re-identification (VI-ReID) needs to be considered in practical applications.

\begin{figure}[t]
    \centering
    \includegraphics[width=1.0\columnwidth]{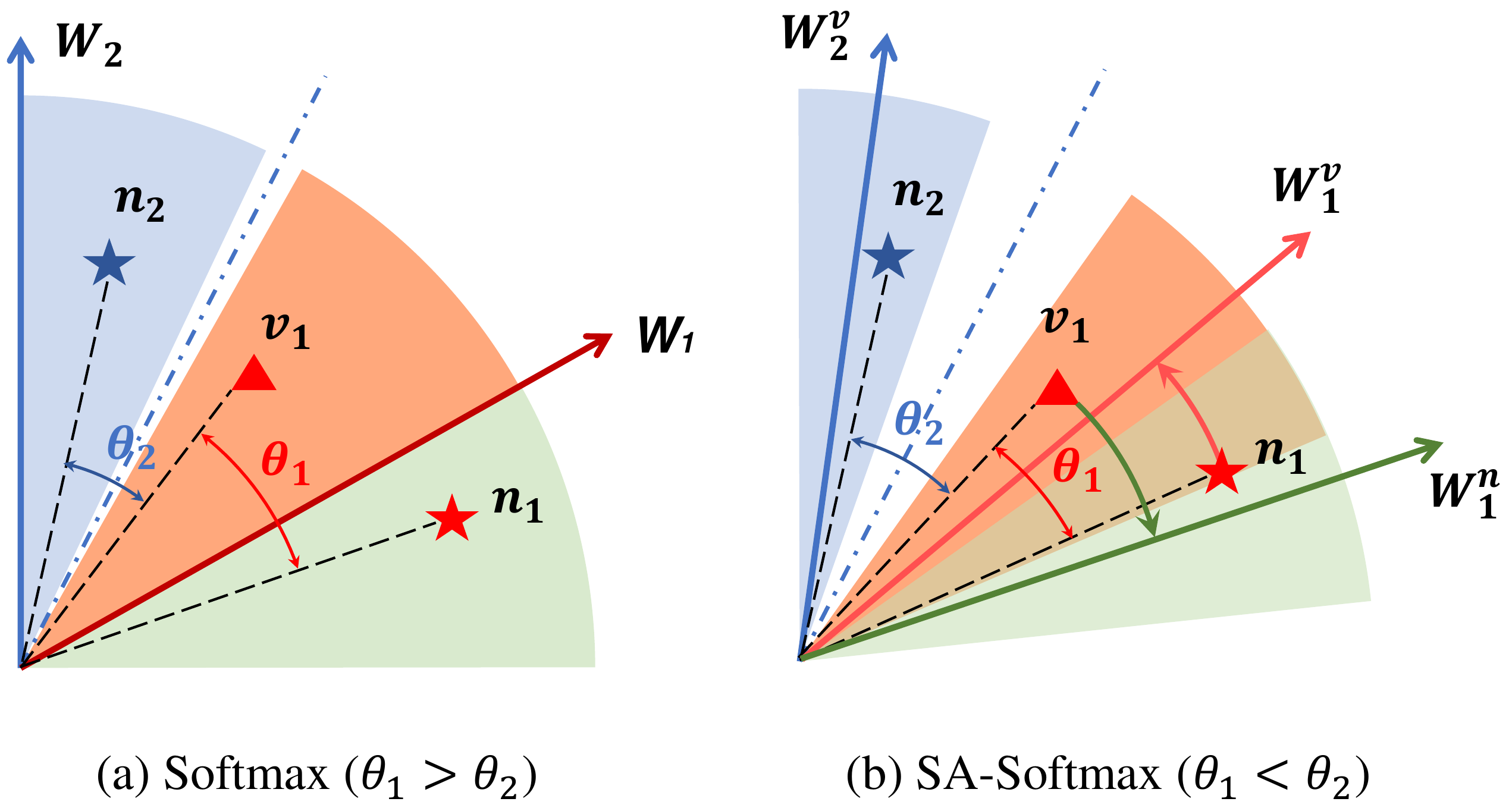}
    \caption{\textbf{Geometrical analysis of the feature space optimized by softmax loss and spectral-aware softmax loss (SA-Softmax).} 
    $v$ and $n$ indicate the visible and near-infrared samples. 
    $W_{1}$ refers to the prototype of class 1. 
    $\theta_1$ and $\theta_2$ refer to the cosine distance of intra-class and inter-class in the retrieval phase, respectively. 
    (a) Since the original softmax loss does not explicitly optimize modality discrepancy, the query feature has to deal with an extra modality bias to match gallery samples from another side (Detailed discussed in Sec. \ref{sec:limitation}). 
    (b) The SA-Softmax optimizes the current samples using a prototype of another spectrum with the same identity, which efficiently explored the entire embedding space in the visible-infrared Re-ID.}
    \label{fig:softmax}
\end{figure}

Compared to the conventional Re-ID task, the primary challenge in such an application lies in the significant modality discrepancy between the VIS and NIR images, which leads to large intra-class variations.
Two typical frameworks are proposed to address this challenge from image generation and representation learning, respectively. 
Image generation approaches~\cite{wang2018learning,wang2019aligngan,wang2019learning} aim to explore stable transformations across the spectrum so that samples under different spectral can be directly compared in the target domain. 
Although with the progress and adaptation of image synthesis methods such as Generative Adversarial Networks (GAN) \cite{goodfellow2014generative,zhu2017unpaired}, this body of works shows impressive visual quality, the synthesized images are still far from photo-realistic. 
Moreover, lacking paired cross-spectral images also makes the training much more challenging. 
Another line of methods is representation learning approaches~\cite{dai2018cross,ye2018visible,ye2018hierarchical,lu2020cross,wu2020rgb}, which aim to bridge the modality gap in the common latent subspace. 
These approaches combine a well-designed loss function with a single-path or multi-path deep neural network. 
Benefiting from end-to-end optimization, these approaches are highly effective and often achieve state-of-the-art performance. 

Despite the substantial progress, we observe that the softmax loss training paradigm, which is designed for a single modality classification task, still be considered as a basic training strategy in the cross-modality retrieval task. 
Even though the softmax loss has made significant progress in single-modality tasks, as shown in Fig.~\ref{fig:softmax} (a), for the cross-spectral condition, the softmax loss considers the VIS and NIR samples from the same identity as a single class in the optimization, ignoring the gap between the two modalities. 
The distributions of the same identity under different spectrums show limited overlapping in the softmax-trained feature space. 
Hence, even after obtaining a well-trained feature distribution, the query feature still has to face an extra modality bias when finding a gallery sample from another side in the retrieval. 
It makes the query feature easily influenced by noise gallery samples and induces a performance penalty. 
Several previous works can make sense in this situation~\cite{schroff2015facenet,wen2016discriminative}, but they either rely on the limited inner-batch relations or only focus on intra-class samples, which do not explore the embedding space in its entirety.

In this paper, we propose a spectral-aware softmax (SA-Softmax) loss to enforce a higher similarity between the cross-modality sample pairs. 
In general, the weight of the final fully connected layer of a deep neural network (DNN) trained with the softmax loss could be considered as a prototype for each identity~\cite{deng2019arcface,deng2021variational}. 
Therefore, the SA-Softmax emphasizes the modality-gap of each modality in every single identity by dividing the prototype of each identity into VIS and NIR prototypes first.
As shown in Fig.~\ref{fig:softmax} (b), to increase the overlapping between the embedding of VIS and NIR distributions, the SA-Softmax utilizes the prototype from another spectrum to optimize the current spectral samples. 
Since the feature embedding and the prototype matrix $W$ have different optimization goals in SA-Softmax, we utilize asynchronous optimizing processing for the feature embedding and the prototype matrix $W$. 
Aside from the metric learning perspective, if regarding the spectrum-aware prototype matrix $W$ in SA-Softmax as a discriminator and the CNN feature extractor as a generator, the SA-Softmax can be considered as an adversarial learning strategy.

While SA-Softmax provides good eventual distribution, it still shows poor optimization for the feature embedding in two perspectives. 
Firstly, although adopting the prototype from another spectrum as the target, the samples in the optimization will also be highly influenced by the prototype from the current spectrum. 
This influence may decrease the similarity of sample pairs in the optimization phase (Detailed discussion can refer to the Sec.~\ref{sec:FM}). 
To overcome this drawback, we employ a feature mask to exclude the response between the sample and the prototype from the current spectrum for better optimization. 
Secondly, we observe that Softmax loss is more concerned with obtaining a high relative results rather than a high similarity for each identity, implying that it has little encouragement to provide a high intra-class similarity (Detailed discussion can be found in Sec.~\ref{sec:AST}). 
Unfortunately, the SA-Softmax inherits this weakness as well. 
Hence, we add an absolute-similarity term in the SA-Softmax to decrease the variation caused by the modality, pose, and occlusion in the VI-ReID and make the distribution of every identity more compact. 

The main contributions of the paper are summarized as follows:
\begin{itemize}
    \item We point out the weakness of the softmax loss training paradigm in the visible-infrared person re-identification problem and propose an asynchronous optimized Spectral-Aware Softmax (SA-Softmax) loss that fully explores the embedding space with the modality information.
    
    \item Combining with the feature mask and the absolute-similarity term, we present an effective framework based on the SA-Softmax for the VI-ReID task.
    
    \item Extensive experiments on two public datasets, RegDB and SYSU-MM01, demonstrate the superiority of our SA-Softmax loss.
\end{itemize}

\section{Related Works}
\textbf{Visible-infrared Person Re-ID.}
On account of the recent advances in deep learning, two typical frameworks are proposed to address this challenge from image generation and representation learning respectively. 
Image generation approaches attempt to build an efficient transformation among the modalities. Wu et al. \cite{wu2017rgb} propose a zero-padding framework to align the cross-modality images in a common space. 
D$^2$RL \cite{wang2019learning} regards each of the samples in a hyperspectral perspective and utilizes the variational autoencoders (VAE) to supply the missing channels for each modality. 
AlignGan \cite{wang2019aligngan} employs a cycle-gan framework with efficient constraints to align from the pixel-level and feature-level. 
Hi-CMD \cite{choi2020hi} uses a GAN framework to change the pose and illumination attributes for each sample to capture ID-discriminative and color-irrelevant representations. 
Ye et al. \cite{ye2021channel}, X-modality \cite{li2020infrared}, and SMCL \cite{ye2018visible} generate an extra modality for jointly learning. On the other hand, representation learning approaches attempt to obtain a modality-irrelevant feature space through module design and constraints. 
From the representation learning perspective, Dai et al. \cite{dai2018cross} employ an adversarial learning strategy to tackle the modality discrepancy. 
DDAG \cite{ye2020dynamic} uses the graph structure to aggregate representations from local to global.  
Wu et al. \cite{wu2021discover} attempt to discover modality-irrelevant nuances in different patterns to match the cross-modality pedestrian samples in a fine-grid way. 
Hao et al. \cite{hao2021cross} combine adversarial learning with center aggregation to extract the centralization features with diversity. 
Despite significant achievements in the VI-ReID task over the past few years, the softmax loss training paradigm designed for a single modality classification task is still considered as a basic training strategy in this area. 
As has been aforementioned, the drawback of softmax loss in such a cross-modality condition is obvious. 

\textbf{Deep Metric Learning.} 
As the optimization target, the loss function plays an essential role in feature representation learning. 
Based on the most widely used Softmax loss, Large Margin Softmax \cite{liu2016large} firstly adds an extra margin to learn a more discriminative feature embedding. 
Since the softmax loss does not explicitly optimize the cosine-similarity, Normface~\cite{wang2017normface} applies the $L2$ normalization to the weight matrix and feature. 
More work is now being done to investigate a better strategy for adjusting the adding margin in the L2-normalized softmax, such as SphereFace~\cite{liu2017sphereface}, CosFace~\cite{wang2018cosface}, and ArcFace~\cite{deng2019arcface}. 
MagFace~\cite{meng2021magface} emphasizes the value of both direction and magnitude for the feature vector and shows the quality of faces through the magnitude of feature embedding. 
While the softmax loss strategy mainly focuses on the cosine similarity, some other works consider the Euclidean feature space as well. 
Triplet loss explores the optimization for pair-wise distance by the triplet relation in each mini-batch. 
Center loss~\cite{wen2016discriminative} adopts the center of each class to obtain a more compact intra-class distribution. 
Circle loss~\cite{sun2020circle} provides a unified perspective for maximizing intra-class similarity while minimizing inter-class similarity.  
Although these metric learning methods perform well on related tasks such as face recognition or person re-identification, they are all designed for single modality tasks that follows the prior that all class samples come from the same domain.
The straight idea that clusters all samples from the same class together ignores the characteristics of multi-modality data. 
Therefore, adopting these learning strategies under cross-modality conditions will face inevitable modality gaps and achieve limited performance.

\section{Method}
\subsection{The limitation of Softmax Loss}
\label{sec:limitation}
Generally, given extracted feature embedding $x_i$, the most widely used loss paradigm in classification, softmax loss, can be formulated as:
\begin{equation}
\begin{split}
\label{eq:orisoftmax}
&\mathcal{L}_{s}(W, x_i) = -log\frac{e^{s_{i}\cos(\theta_{i})}}{\sum^N_{j=1}e^{s_j\cos(\theta_{j})}},\\
& with \quad s_j=\left \|W^{T}_{j} \right \| \left \|x_{i} \right \|, \theta_{j}=\left \langle x_{i}, W_{j} \right \rangle,
\end{split}
\end{equation}
where $W_j \in \mathbb{R}^d$ denotes the $j$-th column of the prototype matrix $W \in \mathbb{R}^{d \times N}$. $d$ is the dimension of embedding feature, $N$ is the number of class, and $x_i \in \mathbb{R}^d$ refers to the $i$-th training sample, which belongs to the $y_i$-th class. 

Eq. \ref{eq:orisoftmax} indicates that the original softmax loss aims to obtain an ideal embedding space by enlarging the intra-class similarity while decreasing the inter-class similarity. 
However, we can also observe several typical drawbacks of the original softmax that limit its performance in cross-spectral conditions.
\textbf{Firstly}, the prototype matrix $W$ is optimized by the whole identity samples containing both two modalities. 
In general, the optimization of softmax loss is considered as two synchronous parts: the prototype matrix $W$ and the embedding feature $x_i$, whose derivatives are described respectively as:
\begin{equation}
\begin{split}
\label{eq:smd}
&\frac{\partial \mathcal{L}_{softmax} }{\partial x_i} = \sum_{j=1}^{N}(p_{ij}-\mathit{1} (j==y_i))W_j,\\
&\frac{\partial \mathcal{L}_{softmax} }{\partial W_i} = \sum_{i=1}^{B}(p_{ij}-\mathit{1} (j==y_i))x_i,\\
&with \qquad p_{ij} = \frac{e^{W^{T}_{y_i}}x_i}{\sum^N_{j=1}e^{W^{T}_{j}x_i}},
\end{split}
\end{equation}
where $y_i$ is the label of $x_i$, and B refers to the batch size.

During the training phase, as indicated in Eq. \ref{eq:smd}, the softmax loss aims to optimize the sample-to-prototype similarity, while the prototype matrix $W$ optimized by the two modality samples is not a suitable target in such a condition. 
Specifically, as shown in Fig. \ref{fig:softmax} (a), for the samples $v_1$ and $n_1$ from the same identity and $n_2$ from another identity, although they have achieved a good training result as $\cos (\left \langle v_{1}, W_{1} \right \rangle) > \cos (\left \langle v_{1}, W_{2} \right \rangle)$, $\cos (\left \langle n_{1}, W_{1} \right \rangle) > \cos (\left \langle n_{1}, W_{2} \right \rangle)$, and $\cos (\left \langle n_{2}, W_{2} \right \rangle) > \cos (\left \langle n_{2}, W_{1} \right \rangle)$, a error retrieval result still occurs as $\cos (\left \langle v_{1}, n_{2} \right \rangle) > \cos (\left \langle v_{1}, n_{1} \right \rangle)$. 

\textbf{Secondly}, the softmax is not a balanced optimization for every identity. 
The first-order derivative and partial derivative of $L(\theta_i)$ are:
\begin{equation}
\begin{split}
\label{eq:ts}
&\mathcal{L}_{s}'(\theta_i)=\frac{e^{s\cos(\theta_j)}s}{e^{s\cos(\theta_i)}+e^{s\cos(\theta_j)}}\sin(\theta_i)>0,\\
&\frac{\partial \mathcal{L}_{s}'(\theta_i)}{\partial \theta_j} =-\frac{e^{s\cos(\theta_i)+s\cos(\theta_j)}s^2\sin(\theta_j)}{(e^{s\cos(\theta_i)}+e^{s\cos(\theta_j)})^2}\sin(\theta_i)<0.\\
\end{split}
\end{equation}

From Eq.\ref{eq:ts}, we can observe that the $\mathcal{L}_{s}'(\theta_i)$ increases monotonically and strictly with the decrease of the $\theta_j$. 
It indicates that those distributions with a high $\theta_j$ at the boundary of the training set are less encouraged to learn compact distributions. 
Since the training set and the testing set come from different identities, the distributions of the two parts are not well aligned. 
Therefore, those distributions at the boundary of the training set also play an important role in learning a more efficient model.

\textbf{Thirdly}, even ignoring the influence of $s$, the optimization is still inefficient. From Eq.\ref{eq:ts}, after $\theta_i < \theta_j$, the $\frac{e^{s\cos(\theta_j)}s}{e^{s\cos(\theta_i)}+e^{s\cos(\theta_j)}}$ in $\mathcal{L}_{s}'(\theta_i)$ will also drop rapidly and show less ability to further decrease the $\theta_i$.

Therefore, we propose a spectral-aware loss to overcome the above drawback of softmax loss in the cross-spectral task.

\begin{figure*}[t]
\centering
\includegraphics[height=8.4cm,width=18cm]{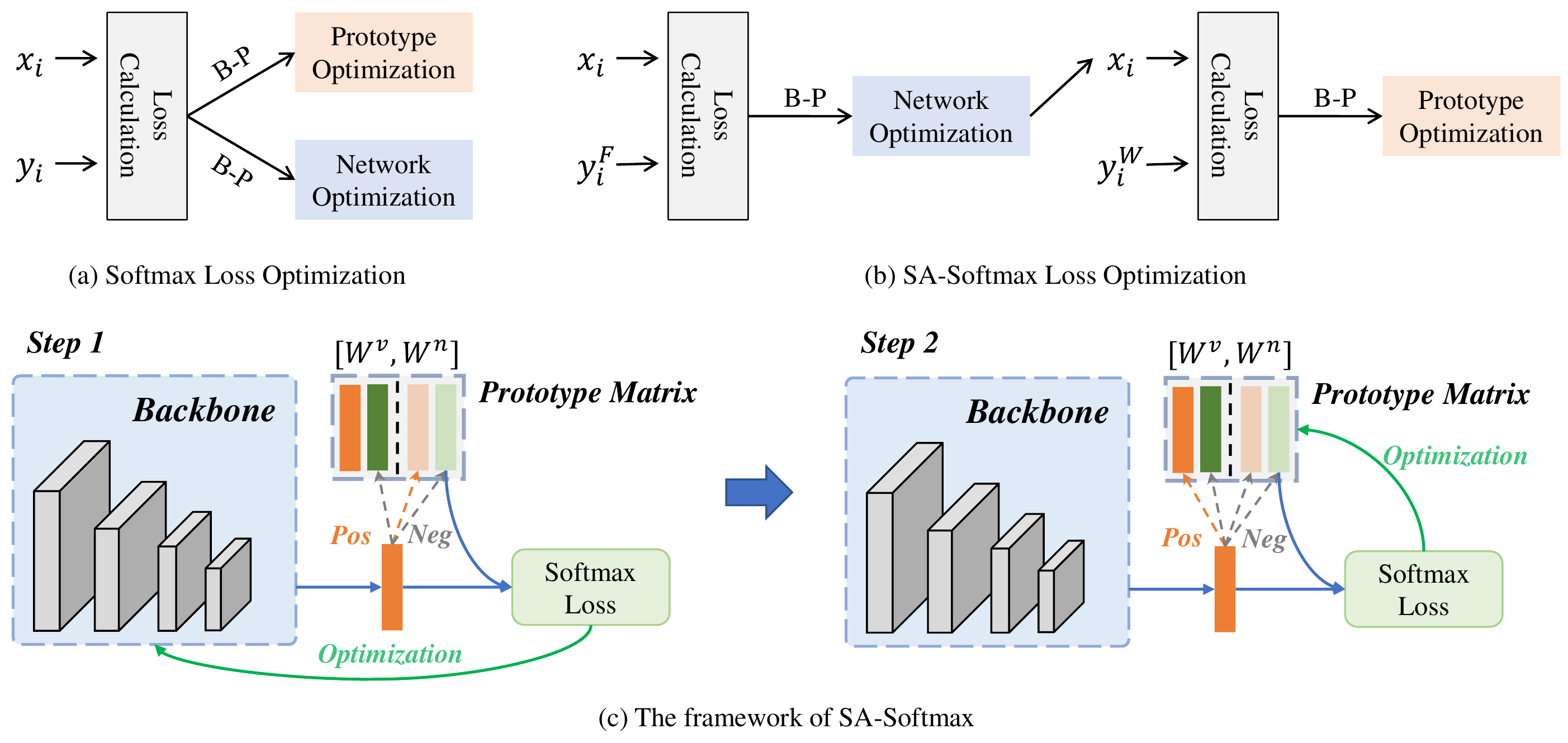}
\vspace{-1em}\caption{\textbf{The optimization process of SA-Softmax loss.} 
(a) The softmax loss uses a synchronization strategy to optimize the network parameters and prototype matrix. 
(b) For SA-Softmax loss, the network and prototype have different goals. 
Hence, SA-Softmax adopts an asynchronous strategy to optimize the network parameters and prototype matrix. 
(c) The two-step asynchronous framework for SA-Softmax loss.}
\label{fig:optim}
\end{figure*}

\subsection{Spectral-Aware Softmax Loss}
Despite its success in single modality tasks, using a prototype to aggregate two modality samples with the same identity lacks explicit optimization for the modality discrepancy and introduces a typical gap in the final feature space, as discussed above. 
Therefore, SA-Softmax inherits the idea of softmax loss and attempts to fully utilize the extra modality information in each sample pair. 

Specifically, as shown in Fig.~\ref{fig:optim}, the SA-Softmax firstly utilizes the modality prototype $[W^v, W^n] \in \mathbb{R}^{d \times 2N}$ instead of the identity prototype $W \in \mathbb{R}^{d \times N}$. Herein, $d$ is the dimension of prototypes, and $N$ is the number of classes. 
Then, to enforce a high overlapping between the distribution of two modalities, the SA-Softmax uses the prototype from another spectrum to optimize each sample. 
To achieve this kind of cross optimization for each sample and corresponding prototype, the SA-Softmax employs an asynchronous optimization strategy for the prototype matrix $[W^v, W^n]$ and sample feature $x_i$, as shown in Fig.~\ref{fig:optim}. 
Therefore, the label $y_i$ of $x_i$ is rewritten to $y^W_i$ and $y^F_i$, respectively, to optimize the prototype matrix and feature extractor as:

\begin{equation}
\begin{split}
\label{eq:saslabel}
\begin{cases}
  & y^W_i = y_i, y^F_i = y_i + N \qquad \text{ if } x_i \in VIS \\
  & y^W_i = y_i + N, y^F_i = y_i \qquad \text{ if } x_i \in NIR.
\end{cases}
\end{split}
\end{equation}

For the prototype matrix $[W^v, W^n]$, this process can be formulated as:
\begin{equation}
\begin{split}
\label{eq:sasw}
&\frac{\partial \mathcal{L}^W_{sas} }{\partial [W^v, W^n]} = \sum_{i=1}^{B}(p_{ij}-\mathit{1} (j==y^W_i))x_i,\\
&with \qquad p_{ij} = \frac{e^{[W^v, W^n]^{T}_{y^W_i}x_i}}{\sum^{2N}_{j=1}e^{[W^v, W^n]^{T}_{j}x_i}}.
\end{split}
\end{equation}
And for feature embedding $x_i$, this process can be formulated as:
\begin{equation}
\begin{split}
\label{eq:sasf}
&\frac{\partial \mathcal{L}^F_{sas} }{\partial x_i} = \sum_{j=1}^{N}(p_{ij}-\mathit{1} (j==y^F_i))[W^v, W^n],\\
&with \qquad p_{ij} = \frac{e^{[W^v, W^n]^{T}_{y^F_i}x_i}}{\sum^{2N}_{j=1}e^{[W^v, W^n]^{T}_{j}x_i}}.
\end{split}
\end{equation}
Here, the $\mathcal{L}^W_{sas}$ and $\mathcal{L}^F_{sas}$ denote the softmax loss results under the label $y^W$ and $y^F$, respectively. 
In summary, the SA-Softmax could be given by:
\begin{equation}
\begin{split}
\label{eq:sas}
&\mathcal{L}_{sas}(W, x_i) = \mathcal{L}^W_{sas}(W) + \mathcal{L}^F_{sas}(x_i),\\
& \mathcal{L}^W_{sas}(W) = -log\frac{e^{[W^v, W^n]^{T}_{y^W_i}x_i}}{\sum^{2N}_{j=1}e^{[W^v, W^n]^{T}_{j}x_i}},\\
& \mathcal{L}^F_{sas}(x_i) = -log\frac{e^{[W^v, W^n]^{T}_{y^F_i}x_i}}{\sum^{2N}_{j=1}e^{[W^v, W^n]^{T}_{j}x_i}}.
\end{split}
\end{equation}

%

\subsection{Feature Mask}
\label{sec:FM}
\begin{figure}[t]
    \centering
    \includegraphics[height=4.5cm,width=8.4cm]{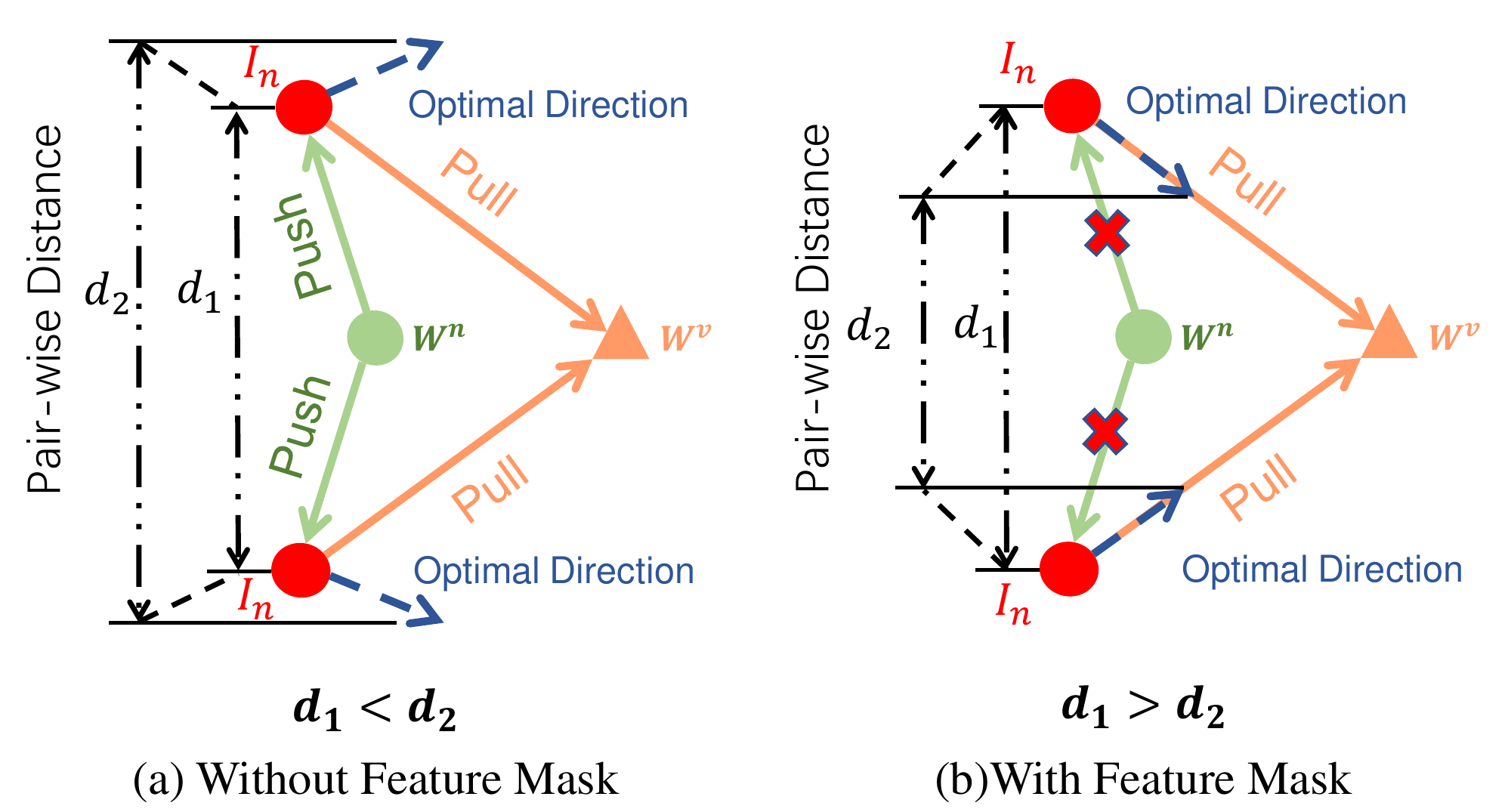}
    \caption{\textbf{A geometrical interpretation of the feature mask.} 
    $d_1$ and $d_2$ refer to the pair-wise distance for same identity before and after the optimization,respectively. 
    (a) Ambiguous optimization in the original SA-Softmax. 
    The prototype of the current modality will push away the training samples, which may lead to ambiguous conditions where the distance between the target prototype and samples decreases (loss decreases) while the discrepancy between the intra-class sample pairs becomes larger. 
    (b) Feature mask ignores the influence from the prototype of the current modality, resulting in more stable training.}
    \label{fig:mask}
\end{figure}
Although the idea of using the prototype from another modality provides a more compact intra-class distribution for the cross-modality samples, the prototype of the current modality of each sample is not insignificant.
As shown in Fig.~\ref{fig:mask} (a), the prototype of the current modality will greatly affect the optimization and may decrease the similarity of pair-wise samples, which would defeat the purpose of SA-Softmax. 
To overcome the tendency, we apply a Feature Mask to eliminate the influence of the current modality prototype for each identity. 
In Fig.~\ref{fig:mask} (b), we show how the feature mask works. 
After combining the feature mask, the $L^F_{sas}$ can be formulated as:
\begin{equation}
\label{eq:sasFM}
\mathcal{L}^F_{sas}(x_i) = -log\frac{e^{[W^v, W^n]^{T}_{y^F_i}x_i}}{\sum^{2N}_{j=1, j \ne y^W_i}e^{[W^v, W^n]^{T}_{j}x_i}}.
\end{equation}

Compared to the original SA-Softmax, adding the feature mask can significantly alleviate the ambiguous optimization, which indicates better learning results. 

\subsection{Absolute-Similarity Term}
\label{sec:AST}
As mentioned above, in Eq.~\ref{eq:orisoftmax} and Eq.~\ref{eq:sas}, we can observe that the softmax loss uses a relative strategy to balance the intra/inter similarity.
It indicates that softmax loss is not a balance optimization. 
Those outlier identities with lower $D_{inter}$ are less likely to learn a more compact intra-class distribution, making them more vulnerable to changes in pose, illumination, and so on. 
On the other hand, owing to its design, even for those with normal identities, the softmax loss shows limited ability to push them to be more similar.
As shown in Eq. \ref{eq:orisoftmax}, the distance $D_{intra}$ can be formulated as:
\begin{equation}
\label{eq:dinter}
D_{intra}(x_i) = e^{W^{T}_{y_i}x_i} = e^{\left \|W_{y_i} \right \| \left \|x_i \right \| cos<W_{y_i}, x_i>},
\end{equation}
where $<W_{y^F_i}, x_i>$ is the angle between $W_{y_i}$ and $x_i$. 
Although it aims to increase the cosine similarity between $W_{y^F_i}$ and $x_i$, from Eq.~\ref{eq:dinter}, the $D_{intra}$ also enjoys the increase of $\left \|W_{y_i} \right \|$ and $\left \|x_i \right \|$, resulting in limited results.
As mentioned in several previous works \cite{wu2020rgb}, since the identities during training and testing do not overlap, the feature has different distributions in the training and testing sets. 
Therefore, if we consider each identity as a single small domain, it is obvious that an ideal compact distribution with less variance is an equal demand for every identity. 

To alleviate this problem, we add an extra Absolute-Similarity Term (AST) to directly enforce that each sample is more similar to its target prototype. 
Following the formulation of the softmax loss, the AST can be given by:
\begin{equation}
\label{eq:ast}
\mathcal{L}_{AST}(x)= \sum_{i=1}^{B}\left \|1-cos<W_{y^F_i}, x_i>  \right \|_2.
\end{equation}

In summary, the final loss function could be shown as:
\begin{equation}
\label{eq:final}
\mathcal{L}
= 
\mathop{\underline{\alpha \mathcal{L}_{sas} + (1-\alpha) \mathcal{L}_{softmax}}}_{Relative\; Term} 
+ 
\mathop{\underline{\beta \mathcal{L}_{AST}}}_{Absolute\; Term},
\end{equation}
where $\alpha$ and $\beta$ are the hyper-parameters to trade-off among each part.

\section{Experiment}
\subsection{Experimental Setting}
\textbf{Datasets.} We conduct experiments on two publicly available visible-infrared person re-identification datasets SYSU-MM01 \cite{wu2017rgb} and RegDB \cite{nguyen2017person}.
\begin{itemize}
    \item \textbf{SYSU-MM01} is a large-scale dataset captured by four visible cameras and two infrared cameras in both indoor and outdoor environments. 
    The training set contains 395 identities with $22,258$ visible images and $11,909$ infrared images, while the testing set includes $96$ identities with $3,803$ infrared images as the query. 
    This dataset contains two different search modes, the all-search mode and the indoor-search mode. In the all-search mode, the gallery images are from all the visible cameras. For the indoor-search mode, the source of the gallery set excludes two outdoor cameras.
    \item \textbf{RegDB} dataset is collected by two aligned cameras, one for visible and the other for far-infrared (thermal). 
    It contains $412$ identities, each with $10$ visible images and $10$ infrared images. Following the evaluation protocol of previous works \cite{ye2018hierarchical,ye2020cross}, we choose half of the identities at random for training and the other half for testing. 
    The results are the average of $10$ repeating.
\end{itemize}

\textbf{Evaluation Protocol.} We follow the evaluation settings in existing VI-ReID methods\cite{wu2021discover,ye2021channel} and adopt the widely used Cumulative Matching Characteristic (CMC) and mean Average Precision (\emph{m}AP) as evaluation metrics.

\textbf{Implementation details.} We use Pytorch to implement our method and finish all the experiments on a single RTX 3090 GPU. 
The mini-batch size is set to 128. 
For each mini-batch, we randomly select 8 identities, each with 8 visible images and 8 infrared images. 
We select the ResNet-50 based PCB \cite{sun2018beyond} model as the baseline, which is a widely used fine-grid part feature learning framework in both Re-ID and visible-infrared Re-ID. 
Following previous works \cite{ye2020dynamic,hao2021cross}, we divide the first convolutional layer to tackle the two modalities input. 
We resize all of the images to 384 $\times$ 192 and use random flipping and random erasing \cite{zhong2020random,ye2021channel} to augment the data. 
To train the model, the SGD optimizer is used with an initial learning rate of 0.01, which is divided by 10 at the 40th and 80th epoch.

\begin{table*}[t]
  \centering
  \renewcommand{\arraystretch}{1.2}
  \caption{\textbf{Comparison with the state-of-the-arts on SYSU-MM01 and RegDB datasets}. R-1, 10, 20 denotes the Rank-1, 10, 20 accuracy.}
  \resizebox{\textwidth}{!}{
  \begin{tabular}{r|c|cccc|cccc|cccc|cccc}
  \hline
   \multirow{3}{*}{Model} & \multirow{3}{*}{Venue} & \multicolumn{8}{c}{RegDB} & \multicolumn{8}{c}{SYSU-MM01} \\
   \cline{3-18}
   & &\multicolumn{4}{c}{Visible to Thermal}&\multicolumn{4}{c}{Thermal to Visible} &\multicolumn{4}{c}{All Search} &\multicolumn{4}{c}{Indoor Search} \\
   \cline{3-18}
   & & R-1 & R-10 & R-20 & \emph{m}AP   & R-1 & R-10 & R-20 & \emph{m}AP & R-1 & R-10 & R-20 & \emph{m}AP & R-1 & R-10 & R-20 & \emph{m}AP\\
   \hline
    Zero-Padding\cite{wu2017rgb}        & ICCV'17   & 17.8 & 34.2 & 44.4 & 18.9 & 16.6 & 34.7 & 44.3 & 17.8 & 14.8 & 54.1 & 71.3 & 15.9 & 20.6 & 68.4 & 85.8 & 26.9 \\
    HCML\cite{ye2018hierarchical}       & AAAI'18   & 24.4 & 47.5 & 56.8 & 20.8 & 21.7 & 45.0 & 55.6 & 22.2 & 14.3 & 53.2 & 69.2 & 16.2 & 24.5 & 73.3 & 86.7 & 30.1 \\
    BDTR\cite{ye2018visible}            & IJCAI'18  & 33.6 & 58.6 & 67.4 & 32.8 & 32.9 & 58.5 & 68.4 & 32.0 & 17.0 & 55.4 & 72.0 & 19.7 & - & - & - & - \\
    cmGAN\cite{dai2018cross}            & IJCAI'18  & -    & -    & -    & -    & -    & -    & -    & -    & 27.0 & 67.5 & 80.6 & 27.8 & 31.7 & 77.2 & 89.2 & 42.2 \\
    AlignGAN\cite{wang2019aligngan}     & ICCV'19   & 57.9 & -    & -    & 53.6 & 56.3 & - & - & 53.4 & 42.4 & 85.0 & 93.7 & 40.7 & 45.9 & 87.6 & 94.4 & 54.3 \\
    X-Modality\cite{li2020infrared}     & AAAI'20   & 62.2 & 83.1 & 91.7 & 60.2 & - & - & - & - & 49.9 & 89.8 & 96.0 & 50.7 & -    & -    & -    & -    \\
    MGE + FMASP\cite{wu2020rgb}         & IJCV'20   & 65.1 & 83.7 & - & 64.5 & - & - & - & - & 43.6 & 86.3 & - & 45.0 & 48.6 & 89.5 & - & 57.5 \\
    SSFT\cite{lu2020cross}             & CVPR'20   & 72.3 & -    & - & 72.9 & 71.0 & - & - & 71.7 & 61.6 & 89.2  & 93.9 & 63.3  & 70.5 & 94.9 & 97.7 & 72.6 \\
    DDAG\cite{ye2020dynamic}            & ECCV'20   & 69.3 & 86.2 & 91.5 & 63.5 & 68.1 & 85.2 & 90.3 & 61.8 & 54.8 & 90.4 & 95.8 & 53.0 & 61.0 & 94.1 & 98.4 & 68.0 \\
    DG-VAE\cite{pu2020dual}             & MM'20     & 73.0 & 86.9 & - & 71.8 & - & - & - & - & 59.5 & 93.8 & - & 58.5 & - & - & - & - \\
    CICL + IAMA\cite{zhao2021joint}     & AAAI'21   & 78.8 & -    & - & 69.4 & 77.9 & - & - & 69.4 & 57.2 & 94.3 & 98.4 & 59.3 & 66.6 & 98.8 & 99.7 & 74.7 \\
    VCD + VML\cite{tian2021farewell}    & CVPR'21   & 73.2 & -    & - & 71.6 & 71.8 & - & - & 70.1 & 60.0 & 94.2 & 98.1 & 58.8 & 66.1 & 96.6 & 99.4 & 73.0 \\
    MPANet\cite{wu2021discover}         & CVPR'21   & 82.8 & -    & - & 80.7 & 83.7 & - & - & 80.9 & 70.6 & 96.2 & 98.8 & 68.2 & 76.7 & 98.2 & 99.6 & 81.0 \\
    MCLNet\cite{hao2021cross}           & ICCV'21   & 80.3 & 92.7 & 96.0 & 73.1 & 75.9 & 90.9 & 94.6 & 69.5 & 65.4 & 93.3 & 97.1 & 62.0 & 72.6 & 97.0 & 99.2 & 76.6 \\
    SMCL\cite{wei2021syncretic}         & ICCV'21   & 83.9 & -    & - & 79.8 & 83.1 & - & - & 78.6 & 67.4 & 92.9 & 96.8 & 61.8 & 68.8 & 96.6 & 98.8 & 75.6 \\
    CAJ\cite{ye2021channel}             & ICCV'21   & 85.0 & 95.5 & 97.5 & 79.1 & 84.8 & 95.3 & 97.5 & 77.8 & 69.9 & 95.7 & 98.5 & 66.9 & 76.3 & 97.9 & 99.5 & 80.4 \\
    \hline
    SA-Softmax                          & \textbf{-} & \textbf{95.2} & \textbf{99.0} & \textbf{99.6} & \textbf{88.7} & \textbf{93.1} & \textbf{98.1} & \textbf{99.1} & \textbf{86.9} & \textbf{71.0} & \textbf{96.9} & \textbf{99.3} & \textbf{68.5} & \textbf{78.8} & \textbf{99.1} & \textbf{99.8} & \textbf{82.2} \\
    \toprule[1pt]
    \end{tabular}}
    \label{Tbl:final}
\end{table*}

\begin{figure*}[t]
\centering
\includegraphics[height=4.5cm,width=18cm]{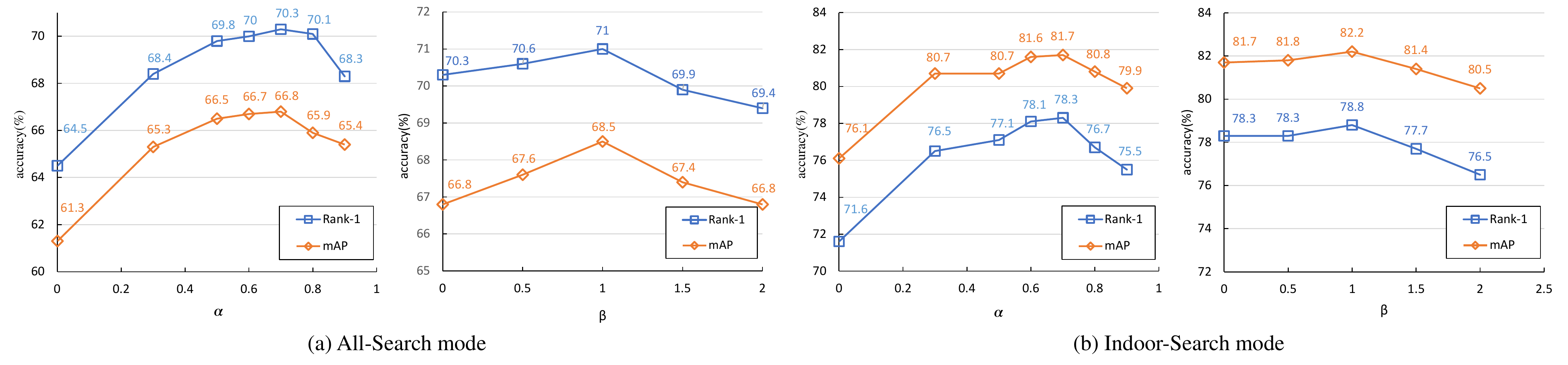}
\vspace{-1em}\caption{\textbf{Impact of the hyper-parameters in terms of CMC (\%) and \emph{m}AP (\%) on SYSU-MM01.} 
For both all-search and indoor-search modes, the performance peaks when  $\alpha$  and $\beta$ are set to 0.7 and 1.0, respectively.}
\label{fig:para}
\end{figure*}

\subsection{Comparison with State-of-the-art Methods}
To comprehensively demonstrate the performance of SA-Softmax, we make a comparison between the combination of the PCB model and SA-Softmax with recent state-of-the-art visible-infrared Re-ID methods in both SYSU-MM01 and RegDB. 
The results are shown in Tab.~\ref{Tbl:final}. 
It is clear that the PCB+SA-Softmax outperforms the existing SOTAs. 
Especially in the RegDB dataset, our method achieves an impressive performance, with Rank-1 accuracy of 95.2\% and \emph{m}AP of 88.7\%  for visible to thermal, and Rank-1 accuracy of 93.1\% and \emph{m}AP of 86.9\% for thermal to visible, respectively. 
Meanwhile, in SYSU-MM01, our method also outperforms the SOTAs in both all-search and indoor-search modes. 
Since the SYSU-MM01 datasets include more divergence in the pose, viewpoints, and background than the RegDB dataset, which focuses primarily on the modality discrepancy, we believe this is why the SA-Softmax achieves fewer improvements in SYSU-MM01 than in RegDB.
Note that SA-Softmax is a plug-and-play module, our network is trained based on the PCB model, and the performance may be boosted with better data augmentations as well as extra well-designed network modules. 

\begin{table}[t]
  \centering
  \renewcommand{\arraystretch}{1.2}
  \caption{\textbf{Ablation study of each components in SA-Softmax in terms of CMC (\%) and \emph{m}AP (\%) on SYSU-MM01.} 'SAS' denotes the original SA-Softamx loss. 'FM' denotes the Feature Mask. 'AST' refers to the Absolute-Similarity Term.}
  \resizebox{85mm}{!}{
  \begin{tabular}{ccc|cc|cc}
  \toprule[1pt]
   \multicolumn{3}{c|}{\textbf{Module}} & \multicolumn{4}{c}{\textbf{SYSU-MM01}} \\
   \cline{1-7}
   \multicolumn{3}{c|}{\emph{Setting}} &\multicolumn{2}{c|}{\emph{All Search}} &\multicolumn{2}{c}{\emph{Indoor Search}}\\
   \cline{1-7}
   SAS & FM & AST & R-1 & \emph{m}AP   & R-1 & \emph{m}AP\\
  \hline
               &           &      & 64.5 & 61.3 & 71.6 & 76.1 \\
    \checkmark &           &      & 69.6 & 66.3 & 77.1 & 81.0 \\
    \checkmark &\checkmark &      & 70.3 & 66.8 & 78.3 & 81.7 \\
    \checkmark &\checkmark &\checkmark      & \textbf{71.0} & \textbf{68.5} & \textbf{78.8} & \textbf{82.2}\\
  \bottomrule[1pt]
    \end{tabular}}
    \label{Tbl:ablation}
\end{table}

\subsection{Ablation Study}
To study the effectiveness of the proposed SA-Softmax loss, we conduct ablation experiments on the baseline PCB model.
In Tab.~\ref{Tbl:ablation}, we show the results of SA-Softmax with different components on the SYSU-MM01. 
Compared to the baseline, the original SA-Softmax greatly improves the performance of SYSU-MM01 in both all-search and indoor-search settings. 
Then, the performance could improve further after adding the feature mask.
Finally, combined with the absolute-similarity terms, the performance achieves Rank-1 accuracy of 71\% and~\emph{m}AP of 68.5\%. 
The results demonstrate that all these components in the SA-Softmax contribute consistently to alleviating modality discrepancy or improving discriminability through an effective training process.

\subsection{Discussion}
\textbf{Impact of the hyper-parameters $\alpha$ and $\beta$.}
In Fig.~\ref{fig:para}, we conduct empirical experiments to measure the model performance under different hyper-parameter settings. 
For the discussion of $\alpha$, we choose the SA-Softmax with Feature Mask as the baseline. 
As shown in the left of Fig.~\ref{fig:para} (a)(b), we observe that the performance increases significantly when attaching the SA-Softmax loss. 
Even if we completely replace the softmax loss with the SA-Softmax loss, the performance under the SYSU-MM01 all-search mode still improves by 3.8\% in Rank-1 accuracy and by 4.1\% in mAP, respectively. 
Besides, the performance peak is reached when $\alpha$ is set to 0.7. 
From the $\beta$ side, as shown in the right of Fig.~\ref{fig:para} (a)(b), after combining the absolute-similarity term (AST) in the training phase, the Rank-1 accuracy and mAP are also improved when $\beta$ is less than 1.5. 
It achieves the best performance when $\beta$ is set to 1.0. 
Since a larger $\beta$ clearly lets makes the network more concerned about encouraging the intra-class similarity, the inter-class similarity will be largely ignored. 
Therefore, the performance degrades when $\beta$ is greater than 1.0. 
Compared to the all-search mode, the AST obtains much smaller increments in the indoor-search mode.
It may be mainly due to the variation among the testing samples. 
Note that the all-search mode includes more diverse backgrounds and poses, which makes it necessary to have a more compact intra-class distribution. 

\begin{table}[t]
  \centering
  \renewcommand{\arraystretch}{1.2}
  \caption{\textbf{Ablation study of weight mask (WM) in SA-Softmax in terms of CMC (\%) and \emph{m}AP (\%) on SYSU-MM01.} Here, we select the SA-Softmax with feature mask (FM) as the baseline.}
  \resizebox{85mm}{!}{
  \begin{tabular}{cc|cc|cc}
  \toprule[1pt]
   \multicolumn{2}{c|}{\textbf{Module}} & \multicolumn{4}{c}{\textbf{SYSU-MM01}} \\
   \cline{1-6}
   \multicolumn{2}{c|}{\emph{Setting}} &\multicolumn{2}{c|}{\emph{All Search}} &\multicolumn{2}{c}{\emph{Indoor Search}}\\
   \cline{1-6}
   SAS + FM & WM & R-1 & \emph{m}AP   & R-1 & \emph{m}AP\\
  \hline
   \checkmark &            & 70.3 & 66.8 & 78.3 & 81.7 \\
   \checkmark & \checkmark & 65.7 & 63.2 & 72.6 & 78.2\\
  \bottomrule[1pt]
    \end{tabular}}
    \label{Tbl:wm}
\end{table}

\begin{figure}[t]
    \centering
    \includegraphics[height=4cm,width=9.0cm]{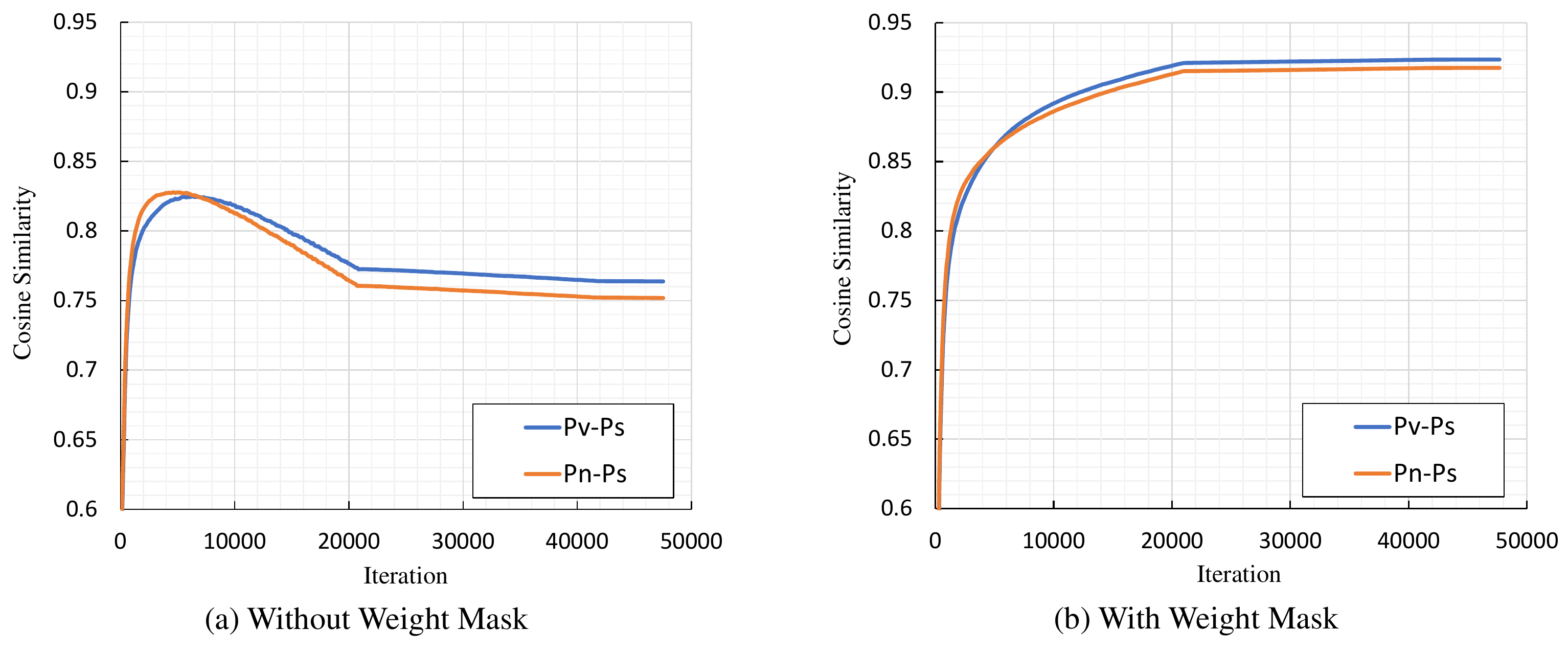}
    \caption{\textbf{The cosine similarity between the modality prototype and the identity prototype with and without the weight mask (WM).} $P_v$ and $P_n$ are prototypes of two modalities in SA-softmax, whereas $P_s$ is the prototype of the original softmax loss. 
    (a) Without WM, the prototype could be supported by another prototype from the same identity. This effect preserves discriminative modality prototypes, thus providing efficient training. 
    (b) Due to the lack of support from another prototype, either $P_v$ or $P_n$ becomes more and more similar to $P_s$, causing the SA-Softmax loss to degrade to the original softmax loss.}
    \label{fig:wm}
\end{figure}

\textbf{Why not add the weight mask?}
Besides the feature mask, the same worry occurs in the optimization for the prototype matrix. 
However, this condition is totally different from the optimization for feature embedding. 
Concretely, if following the above strategy and adding a weight mask, the $L^W_{sas}$ will be formulated as:
\begin{equation}
\begin{split}
\label{eq:sasWM}
&\mathcal{L}^W_{sas}(W) = -log\frac{e^{[W^v, W^n]^{T}_{y^W_i}x_i}}{e^{[W^v, W^n]^{T}_{y^W_i}x_i}+D_{inter}(x_i)},\\
&with \quad D_{inter}(x_i) = {\textstyle \sum^{2N}_{j=1, j \ne y^W_i, j \ne y^F_i}e^{[W^v, W^n]^{T}_{j}x_i}},
\end{split}
\end{equation}
where we could approximate consider the $D_{inter}$ to the inter-class distance in the same part of softmax loss. 
We can easily observe that $L^W_{sas}$ will degrade to the original softmax loss. 
Although the distribution bias from the training set brings some variances between $W^v$ and $W^n$, it provides limited improvement for this task. 
For example, in Fig.~\ref{fig:wm}, we visualize the cosine similarity between the two modality prototypes with the identity prototype (trained by original softmax loss), with and without the weight mask. 
After adding the weight mask, the paired prototypes can support each other. This kind of support can help each prototype leave away from the identity prototype to the decision boundary. It not only avoids the degradation from the modality prototype to the identity prototype but also keeps the effectiveness of the SA-Softmax.
Furthermore, we conduct a quantitative experiment in the SYSU-MM01 to show the influence of the weight mask.  
As shown in Table~\ref{Tbl:wm}, after adding the weight mask, the performance decreased in both rank-1 accuracy and \emph{m}AP. 
Although benefitting from the distribution bias of different modalities in each class, the weight mask still leads to performance degradation.

\begin{table}[t]
  \centering
  \renewcommand{\arraystretch}{1.2}
  \caption{\textbf{Comparison of the SA-Softmax and other softmax-based training strategies on the SYSU-MM01 in terms of CMC (\%) and \emph{m}AP (\%).}}
  \resizebox{85mm}{!}{
  \begin{tabular}{l|cc|cc|cc}
  \toprule[1pt]
   \multirow{3}{*}{\textbf{Method}} & \multicolumn{2}{c|}{\textbf{Hyper-parameter}} & \multicolumn{4}{c}{\textbf{SYSU-MM01}} \\
   \cline{2-7}
   & \multicolumn{2}{c|}{\emph{Setting}} &\multicolumn{2}{c}{\emph{All Search}} & \multicolumn{2}{c}{\emph{Indoor Search}}\\
   \cline{2-7}
   & margin(m) & gamma($\gamma$) & R-1 & \emph{m}AP   & R-1 & \emph{m}AP\\
  \hline
    baseline (Softmax)                       & - & - & 64.5 & 61.3 & 71.6 & 76.1 \\
    AM-Softmax \cite{wang2018additive}       & 0.1 & - & 56.4 & 57.5 & 65.9 & 73.1 \\
    AM-Softmax \cite{wang2018additive}       & 0.2 & - & 55.0 & 55.9 & 63.9 & 71.0 \\
    AM-Softmax \cite{wang2018additive}       & 0.3 & - & 45.2 & 47.5 & 49.5 & 59.7 \\
    Circle Loss \cite{sun2020circle}         & - & 32 & 66.7 & 64.5 & 74.9 & 79.6\\
    Circle Loss \cite{sun2020circle}         & - & 64 & 64.6 & 62.5 & 71.4 & 76.5\\
    Circle Loss \cite{sun2020circle}         & - & 128 & 55.2 & 53.5 & 62.5 & 69.2 \\
    SA-Softmax                               & - & - & \textbf{71.0} & \textbf{68.5} & \textbf{78.8} & \textbf{82.2}\\
  \bottomrule[1pt]
    \end{tabular}}
    \label{Tbl:compare}
\end{table}

\textbf{Effect of SA-Softmax loss.} As we have introduced above, the SA-Softmax loss is not the first one to explore more specific softmax-based loss functions. 
Hence, in this part, we compare SA-Softmax with two typical metric learning strategies: AM-Softmax \cite{wang2018additive} and Circle Loss \cite{sun2020circle} in the SYSU-MM01 dataset.

For a fair comparison, we use the same implementation detail as training the SA-Softmax. 
For the AM-Softmax, we follow the setting provided by~\cite{wei2018person}, in which $m$ and $s$ are set to 0.3 and 15, respectively. 
But we further observe that this hyper-parameter does not work well in the VI-ReID task, as a larger margin $m$ destroys the convergence in the early epochs. 
Therefore, we evaluate the AM-Softmax with different $m$ of 0.3, 0.2, and 0.1, and show all the results. 
Here, we measure the circle loss under the same conditions as in~\cite{sun2020circle}, but with the gamma ranging from 32 to 128 to demonstrate its performance in the VI-ReID task.

The final results are reported in Tab.~\ref{Tbl:compare}.
We can observe that those well-designed deep metric learning methods face completely different application scenarios compared to the single modality retrieval task.
We have to relax the hyper-parameters to ensure they still work.
Interestingly, adding an extra margin may not make sense in a multi-modality condition, as it would make it difficult for the network to converge in the early epochs.
Meanwhile, the SA-Softmax shows significant improvement on the VI-ReID task compared to other deep metric learning strategies.
Specifically, the SA-Softmax outperforms the circle loss by +4.3\% in Rank-1 accuracy and +4.0\% in the \emph{m}AP under the SYSU-MM01 all-search mode.
Besides, in the indoor-search mode, the SA-Softmax also surpasses the circle loss by 3.9\% in Rank-1 accuracy and 2.6\% in \emph{m}AP.
Benefiting from the design which directly emphasizes the modality discrepancy in the VI-ReID task, the SA-Softmax reaches a considerable superiority against other deep metric methods. 
As a plug-and-play module, it indicates the great potential of SA-Softmax to become a new benchmark for the VI-ReID task.

\begin{figure*}[t!]
\centering
\includegraphics[height=18cm,width=18cm]{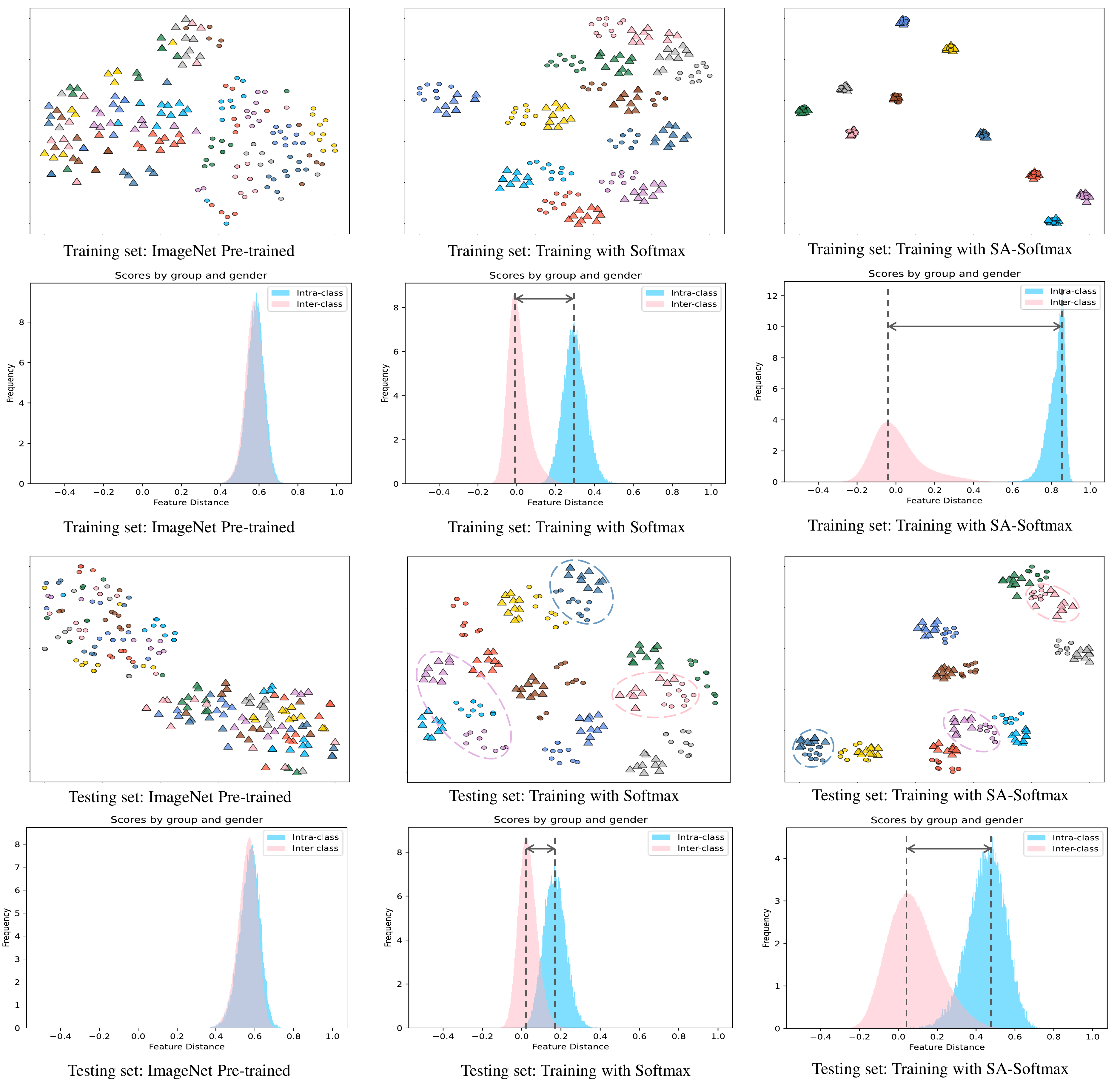}
\vspace{-1em}\caption{\textbf{Visualization of learned features space ($1_{st}$, $3_{rd}$ row) and feature distances histogram ($2_{nd}$, $4_{th}$ row) under different training strategies.} 
The circles and triangles in the features space denote the features extracted from the visible and infrared modalities, respectively.
To better evaluate differences, the similarity between pairs of samples from the same modality has been ignored in the feature distance histograms. 
It is obvious that the SA-Softmax loss utilizes the training set better and obtains a more compact distribution than the original softmax loss.}
\label{fig:tsne}
\end{figure*}

\textbf{Visualization of SA-Softmax loss.} To demonstrate the effectiveness of SA-Softmax, we visualize the well-learned feature spaces in the training and test sets of SYSU-MM01 via t-SNE~\cite{van2008visualizing}. 
For comparison, we also visualize the feature space training by the original softmax loss.
The visualization results are shown in Fig.~\ref{fig:tsne}. 
From the top column of Fig.~\ref{fig:tsne}, we can easily observe the significant modality discrepancy between the two spectral image samples, which is the main challenge of the VI-ReID task. 
The middle column shows the feature space trained by the original softmax loss.
As we have previously discussed, employing the class prototype for cross-modality samples lacks the penalty for the modality discrepancy. 
Therefore, the distributions in each class still show limited overlap between the two spectral samples in the training set.
Besides the limited optimization of intra-class similarity, the lack of a compact distribution also results in class prototypes that do not represent the distribution of each class well, which in turn limits the optimization of the inter-class distribution.
Such inexplicit optimizations in the training set can also lead to unsatisfactory performance in the testing set.

In contrast, as shown in the third column of Fig.~\ref{fig:tsne}, the SA-Softmax can be better adapted to the VI-ReID task, with a more specific design for the cross-modality retrieval task. 
Due to the use of modality prototypes and the explicit optimization direction for the modality gap, the prototypes can better describe the distribution of each class, enabling a more efficient training process.
In the visualized feature space of the training set, SA-Softmax exhibits an impressive performance in exploiting the entire training set, which in turn leads to significant improvements in the test set.

Furthermore, to quantitatively analyze the original softmax and SA-Softmax, we visualized the cosine distance distributions of intra-class cross-modality pairs and inter-class cross-modality sample pairs in Fig.~\ref{fig:tsne}.  
Compared with the softmax loss, the SA-Softmax significantly increases the similarity between intra-class cross-modality pairs and broadens the distribution discrepancy in both training and testing sets. 
In general, the visualization results further demonstrate the superior performance of the SA-Softmax in VI-ReID.

\section{Conclusion}
In this paper, we highlight the limitation of the softmax loss paradigm in the visible-infrared person re-identification task: it ignores the optimization of modality discrepancy. %
To this end, we propose a novel Spectral-Aware Softmax loss (SA-Softmax), which exploits the global-wise modality information and emphasizes the modality discrepancy by employing the modality prototype with an asynchronous optimization. 
Based on the analysis and observation of the ambiguous optimization in the SA–Softmax loss, we modify the SA–Softmax loss with the feature mask and absolute-similarity term to obtain a more stable result. 
By combining the widely-used backbone, the SA-Softmax demonstrates its superiority among softmax-based training strategies and achieves state-of-the-art performance in two publicly available VI-ReID datasets, RegDB, and SYSU-MM01. 

\section*{Acknowledgments}
This work was supported by the National Science Fund for Distinguished Young Scholars (No.62025603), the National Natural Science Foundation of China (No.U1705262, No. 62176222, No. 62176223, No. 62176226, No. 62072386, No. 62072387, No. 62072389, No. 62002305, No. 61772443, No. 61802324 and No. 61702136), Guangdong Basic and Applied Basic Research Foundation (No.2019B1515120049), the Natural Science Foundation of Fujian Province of China (No.2021J01002), and the Fundamental Research Funds for the central universities (No. 20720200077, No. 20720200090 and No. 20720200091).


\bibliographystyle{IEEEtran}
\bibliography{IEEEfull}

\begin{thebibliography}{10}
\providecommand{\url}[1]{#1}
\csname url@samestyle\endcsname
\providecommand{\newblock}{\relax}
\providecommand{\bibinfo}[2]{#2}
\providecommand{\BIBentrySTDinterwordspacing}{\spaceskip=0pt\relax}
\providecommand{\BIBentryALTinterwordstretchfactor}{4}
\providecommand{\BIBentryALTinterwordspacing}{\spaceskip=\fontdimen2\font plus
\BIBentryALTinterwordstretchfactor\fontdimen3\font minus
  \fontdimen4\font\relax}
\providecommand{\BIBforeignlanguage}[2]{{%
\expandafter\ifx\csname l@#1\endcsname\relax
\typeout{** WARNING: IEEEtran.bst: No hyphenation pattern has been}%
\typeout{** loaded for the language `#1'. Using the pattern for}%
\typeout{** the default language instead.}%
\else
\language=\csname l@#1\endcsname
\fi
#2}}
\providecommand{\BIBdecl}{\relax}
\BIBdecl

\bibitem{eom2019learning}
C.~Eom and B.~Ham, ``Learning disentangled representation for robust person
  re-identification,'' in \emph{Proceedings of the NeurIPS}, 2019, pp.
  5297--5308.

\bibitem{Zhai2020ad}
Y.~Zhai, S.~Lu, Q.~Ye, X.~Shan, J.~Chen, R.~Ji, and Y.~Tian, ``Ad-cluster:
  Augmented discriminative clustering for domain adaptive person
  re-identification,'' in \emph{Proceedings of the CVPR}, June 2020.

\bibitem{zheng2019pyramidal}
F.~Zheng, C.~Deng, X.~Sun, X.~Jiang, X.~Guo, Z.~Yu, F.~Huang, and R.~Ji,
  ``Pyramidal person re-identification via multi-loss dynamic training,'' in
  \emph{Proceedings of the CVPR}, 2019, pp. 8514--8522.

\bibitem{zheng2019joint}
Z.~Zheng, X.~Yang, Z.~Yu, L.~Zheng, Y.~Yang, and J.~Kautz, ``Joint
  discriminative and generative learning for person re-identification,'' in
  \emph{Proceedings of the CVPR}, 2019, pp. 2138--2147.

\bibitem{wang2018learning}
G.~Wang, Y.~Yuan, X.~Chen, J.~Li, and X.~Zhou, ``Learning discriminative
  features with multiple granularities for person re-identification,'' in
  \emph{Proceedings of the ACM MM}, 2018, pp. 274--282.

\bibitem{wang2019aligngan}
G.~Wang, T.~Zhang, J.~Cheng, S.~Liu, Y.~Yang, and Z.~Hou, ``Rgb-infrared
  cross-modality person re-identification via joint pixel and feature
  alignment,'' in \emph{Proceedings of the ICCV}, 2019, pp. 3623--3632.

\bibitem{wang2019learning}
Z.~Wang, Z.~Wang, Y.~Zheng, Y.-Y. Chuang, and S.~Satoh, ``Learning to reduce
  dual-level discrepancy for infrared-visible person re-identification,'' in
  \emph{Proceedings of the CVPR}, 2019, pp. 618--626.

\bibitem{goodfellow2014generative}
I.~Goodfellow, J.~Pouget-Abadie, M.~Mirza, B.~Xu, D.~Warde-Farley, S.~Ozair,
  A.~Courville, and Y.~Bengio, ``Generative adversarial nets,'' in
  \emph{Proceedings of the NeurIPS}, 2014, pp. 2672--2680.

\bibitem{zhu2017unpaired}
J.-Y. Zhu, T.~Park, P.~Isola, and A.~A. Efros, ``Unpaired image-to-image
  translation using cycle-consistent adversarial networks,'' in
  \emph{Proceedings of the ICCV}, 2017, pp. 2223--2232.

\bibitem{dai2018cross}
P.~Dai, R.~Ji, H.~Wang, Q.~Wu, and Y.~Huang, ``Cross-modality person
  re-identification with generative adversarial training.'' in
  \emph{Proceedings of the IJCAI}, 2018, pp. 677--683.

\bibitem{ye2018visible}
M.~Ye, Z.~Wang, X.~Lan, and P.~C. Yuen, ``Visible thermal person
  re-identification via dual-constrained top-ranking.'' in \emph{Proceedings of
  the IJCAI}, 2018, pp. 1092--1099.

\bibitem{ye2018hierarchical}
M.~Ye, X.~Lan, J.~Li, and P.~C. Yuen, ``Hierarchical discriminative learning
  for visible thermal person re-identification,'' in \emph{Proceedings of the
  AAAI}, 2018.

\bibitem{lu2020cross}
Y.~Lu, Y.~Wu, B.~Liu, T.~Zhang, B.~Li, Q.~Chu, and N.~Yu, ``Cross-modality
  person re-identification with shared-specific feature transfer,'' in
  \emph{Proceedings of the CVPR}, June 2020.

\bibitem{wu2020rgb}
A.~Wu, W.-S. Zheng, S.~Gong, and J.~Lai, ``Rgb-ir person re-identification by
  cross-modality similarity preservation,'' \emph{IJCV}, pp. 1--21, 2020.

\bibitem{schroff2015facenet}
F.~Schroff, D.~Kalenichenko, and J.~Philbin, ``Facenet: A unified embedding for
  face recognition and clustering,'' in \emph{Proceedings of the CVPR}, 2015,
  pp. 815--823.

\bibitem{wen2016discriminative}
Y.~Wen, K.~Zhang, Z.~Li, and Y.~Qiao, ``A discriminative feature learning
  approach for deep face recognition,'' in \emph{Proceedings of the
  ECCV}.\hskip 1em plus 0.5em minus 0.4em\relax Springer, 2016, pp. 499--515.

\bibitem{deng2019arcface}
J.~Deng, J.~Guo, N.~Xue, and S.~Zafeiriou, ``Arcface: Additive angular margin
  loss for deep face recognition,'' in \emph{Proceedings of the IEEE/CVF
  Conference on Computer Vision and Pattern Recognition}, 2019, pp. 4690--4699.

\bibitem{deng2021variational}
J.~Deng, J.~Guo, J.~Yang, A.~Lattas, and S.~Zafeiriou, ``Variational prototype
  learning for deep face recognition,'' in \emph{Proceedings of the IEEE/CVF
  Conference on Computer Vision and Pattern Recognition}, 2021, pp.
  11\,906--11\,915.

\bibitem{wu2017rgb}
A.~Wu, W.-S. Zheng, H.-X. Yu, S.~Gong, and J.~Lai, ``Rgb-infrared
  cross-modality person re-identification,'' in \emph{Proceedings of the ICCV},
  2017, pp. 5380--5389.

\bibitem{choi2020hi}
S.~Choi, S.~Lee, Y.~Kim, T.~Kim, and C.~Kim, ``Hi-cmd: Hierarchical
  cross-modality disentanglement for visible-infrared person
  re-identification,'' in \emph{Proceedings of the CVPR}, 2020, pp.
  10\,257--10\,266.

\bibitem{ye2021channel}
M.~Ye, W.~Ruan, B.~Du, and M.~Z. Shou, ``Channel augmented joint learning for
  visible-infrared recognition,'' in \emph{Proceedings of the IEEE/CVF
  International Conference on Computer Vision}, 2021, pp. 13\,567--13\,576.

\bibitem{li2020infrared}
D.~Li, X.~Wei, X.~Hong, and Y.~Gong, ``Infrared-visible cross-modal person
  re-identification with an x modality.'' in \emph{Proceedings of the AAAI},
  2020, pp. 4610--4617.

\bibitem{ye2020dynamic}
M.~Ye, J.~Shen, D.~J. Crandall, L.~Shao, and J.~Luo, ``Dynamic dual-attentive
  aggregation learning for visible-infrared person re-identification,'' in
  \emph{Proceedings of the ECCV}, 2020.

\bibitem{wu2021discover}
Q.~Wu, P.~Dai, J.~Chen, C.-W. Lin, Y.~Wu, F.~Huang, B.~Zhong, and R.~Ji,
  ``Discover cross-modality nuances for visible-infrared person
  re-identification,'' in \emph{Proceedings of the IEEE/CVF Conference on
  Computer Vision and Pattern Recognition}, 2021, pp. 4330--4339.

\bibitem{hao2021cross}
X.~Hao, S.~Zhao, M.~Ye, and J.~Shen, ``Cross-modality person re-identification
  via modality confusion and center aggregation,'' in \emph{Proceedings of the
  IEEE/CVF International Conference on Computer Vision}, 2021, pp.
  16\,403--16\,412.

\bibitem{liu2016large}
W.~Liu, Y.~Wen, Z.~Yu, and M.~Yang, ``Large-margin softmax loss for
  convolutional neural networks.'' in \emph{ICML}, vol.~2, no.~3, 2016, p.~7.

\bibitem{wang2017normface}
F.~Wang, X.~Xiang, J.~Cheng, and A.~L. Yuille, ``Normface: L2 hypersphere
  embedding for face verification,'' in \emph{Proceedings of the 25th ACM
  international conference on Multimedia}, 2017, pp. 1041--1049.

\bibitem{liu2017sphereface}
W.~Liu, Y.~Wen, Z.~Yu, M.~Li, B.~Raj, and L.~Song, ``Sphereface: Deep
  hypersphere embedding for face recognition,'' in \emph{Proceedings of the
  IEEE conference on computer vision and pattern recognition}, 2017, pp.
  212--220.

\bibitem{wang2018cosface}
H.~Wang, Y.~Wang, Z.~Zhou, X.~Ji, D.~Gong, J.~Zhou, Z.~Li, and W.~Liu,
  ``Cosface: Large margin cosine loss for deep face recognition,'' in
  \emph{Proceedings of the IEEE conference on computer vision and pattern
  recognition}, 2018, pp. 5265--5274.

\bibitem{meng2021magface}
Q.~Meng, S.~Zhao, Z.~Huang, and F.~Zhou, ``Magface: A universal representation
  for face recognition and quality assessment,'' in \emph{Proceedings of the
  IEEE/CVF Conference on Computer Vision and Pattern Recognition}, 2021, pp.
  14\,225--14\,234.

\bibitem{sun2020circle}
Y.~Sun, C.~Cheng, Y.~Zhang, C.~Zhang, L.~Zheng, Z.~Wang, and Y.~Wei, ``Circle
  loss: A unified perspective of pair similarity optimization,'' in
  \emph{Proceedings of the IEEE/CVF Conference on Computer Vision and Pattern
  Recognition}, 2020, pp. 6398--6407.

\bibitem{nguyen2017person}
D.~T. Nguyen, H.~G. Hong, K.~W. Kim, and K.~R. Park, ``Person recognition
  system based on a combination of body images from visible light and thermal
  cameras,'' \emph{Sensors}, vol.~17, no.~3, p. 605, 2017.

\bibitem{ye2020cross}
M.~Ye, X.~Lan, Q.~Leng, and J.~Shen, ``Cross-modality person re-identification
  via modality-aware collaborative ensemble learning,'' \emph{IEEE Transactions
  on Image Processing}, 2020.

\bibitem{sun2018beyond}
Y.~Sun, L.~Zheng, Y.~Yang, Q.~Tian, and S.~Wang, ``Beyond part models: Person
  retrieval with refined part pooling (and a strong convolutional baseline),''
  in \emph{Proceedings of the ECCV}, 2018, pp. 480--496.

\bibitem{zhong2020random}
Z.~Zhong, L.~Zheng, G.~Kang, S.~Li, and Y.~Yang, ``Random erasing data
  augmentation.'' in \emph{Proceedings of the AAAI}, 2020.

\bibitem{pu2020dual}
N.~Pu, W.~Chen, Y.~Liu, E.~M. Bakker, and M.~S. Lew, ``Dual gaussian-based
  variational subspace disentanglement for visible-infrared person
  re-identification,'' in \emph{Proceedings of the ACM Multimedia}, 2020.

\bibitem{zhao2021joint}
Z.~Zhao, B.~Liu, Q.~Chu, Y.~Lu, and N.~Yu, ``Joint color-irrelevant consistency
  learning and identity-aware modality adaptation for visible-infrared cross
  modality person re-identification,'' in \emph{Proceedings of the AAAI
  Conference on Artificial Intelligence}, vol.~35, no.~4, 2021, pp. 3520--3528.

\bibitem{tian2021farewell}
X.~Tian, Z.~Zhang, S.~Lin, Y.~Qu, Y.~Xie, and L.~Ma, ``Farewell to mutual
  information: Variational distillation for cross-modal person
  re-identification,'' in \emph{Proceedings of the IEEE/CVF Conference on
  Computer Vision and Pattern Recognition}, 2021, pp. 1522--1531.

\bibitem{wei2021syncretic}
Z.~Wei, X.~Yang, N.~Wang, and X.~Gao, ``Syncretic modality collaborative
  learning for visible infrared person re-identification,'' in
  \emph{Proceedings of the IEEE/CVF International Conference on Computer
  Vision}, 2021, pp. 225--234.

\bibitem{wang2018additive}
F.~Wang, J.~Cheng, W.~Liu, and H.~Liu, ``Additive margin softmax for face
  verification,'' \emph{IEEE Signal Processing Letters}, vol.~25, no.~7, pp.
  926--930, 2018.

\bibitem{wei2018person}
L.~Wei, S.~Zhang, W.~Gao, and Q.~Tian, ``Person transfer gan to bridge domain
  gap for person re-identification,'' in \emph{Proceedings of the CVPR}, 2018,
  pp. 79--88.

\bibitem{van2008visualizing}
L.~Van~der Maaten and G.~Hinton, ``Visualizing data using t-sne.''
  \emph{Journal of machine learning research}, vol.~9, no.~11, 2008.

\end{thebibliography}

\newpage

 




\vfill

\end{document}